%% file: main.tex
\documentclass[runningheads]{llncs}
\usepackage{graphicx}

\usepackage{tikz}
\usepackage{comment}
\usepackage{amsmath,amssymb} 
\usepackage{color}

\usepackage{graphicx}
\usepackage{booktabs}
\usepackage[utf8]{inputenc} 
\usepackage[T1]{fontenc}    
\usepackage{url}            
\usepackage{booktabs}       
\usepackage{amsfonts}       
\usepackage{nicefrac}       
\usepackage{microtype}      
\usepackage{xcolor}         
\usepackage{subcaption}
\usepackage{multirow}
\usepackage{multicol}
\usepackage{diagbox}
\usepackage{wrapfig}

\usepackage[accsupp]{axessibility}  

\newcommand{\todo}[1]{}

\newcommand{\tb}[1]{}
\newcommand{\mb}[1]{}
\newcommand{\sg}[1]{}

\newcommand{\modelname}{MOoSe}
\newcommand{\deepens}{DeepEns}
\newcommand{\multiheadens}{MH-Ens}
\newcommand{\globalhead }{Global}

\begin{document}
\pagestyle{headings}
\mainmatter
\def\ECCVSubNumber{7}  

\title{Probing Contextual Diversity for Dense Out-of-Distribution Detection} 

\titlerunning{}
%
\author{Silvio Galesso\and
Maria Alejandra Bravo\and
Mehdi Naouar\and
Thomas Brox}
\authorrunning{}
%
\institute{University of Freiburg 
}
\maketitle

\begin{abstract}
Detection of out-of-distribution (OoD) samples in the context of image classification has recently become an area of interest and active study, along with the topic of uncertainty estimation, to which it is closely related. In this paper we explore the task of OoD segmentation, which has been studied less than its classification counterpart and presents additional challenges. Segmentation is a dense prediction task for which the model's outcome for each pixel depends on its surroundings.
The receptive field and the reliance on context play a role for distinguishing different classes and, correspondingly, for spotting OoD entities. We introduce MOoSe, an efficient strategy to leverage the various levels of context represented within semantic segmentation models and show that even a simple aggregation of multi-scale representations has consistently positive effects on OoD detection and uncertainty estimation.
\end{abstract}

\section{Introduction}
\input{sections/1_introduction}

\section{Related Work}
\input{sections/2_related_work}

\section{Multi-Head Context Networks}
\input{sections/3_method}

\section{Experiments}
\input{sections/4_main_results}


\section{Analysis}
\input{sections/5_analysis_new}

\section{Conclusion}
\input{sections/6_conclusion}

\clearpage
%
%
\bibliographystyle{splncs04}
\bibliography{egbib}

\end{document}

%% file: sections/1_introduction.tex

\input{figures/drawings/teaser}

Imagine you see a pattern, an object, or a scene configuration you do not know. You will identify it as novel and it will attract your attention. This ability to deal with an open world and to identify novel patterns at all semantic levels is one of the many ways how human perception differs from contemporary machine learning. Most deep learning setups assume a closed world with a fixed set of known classes to choose from. However, many real-world tasks do not match this assumption. Very often, maximum deviations from the training samples are the most interesting data points. 

Accordingly, novelty/anomaly/out-of-distribution detection has attracted more and more interest recently. Outside of data regimes with limited variation, such as in industrial inspection~\cite{cohen2020,rudolph2021same}, the common approaches to identify unseen patterns derive uncertainty estimates from an existing classification model and mark samples with large uncertainty as novel or out-of-distribution~\cite{liang2018enhancing,hendrycks2019selfsupervised,bergman2019classification,hein2019relu,winkens2020contrastive}. 
This approach comes with a conflict between the classifier focusing on features that help discriminate between the known classes and the need for rich and diverse features that can identify out-of-distribution patterns. 
This is especially true for semantic segmentation, where a pixel's class is not only defined by its own appearance, but also by the context it appears in.
Based on context information only and ignoring appearance, a segmentation model could assume that a large animal (or an oversized telephone, like in the example in Figure~\ref{fig:teaser}) in the middle of the road is a vehicle, while based on local appearance only it could believe that pictures on a billboard are, for example, actual people in the flesh.

In order to combine context and local appearance, modern segmentation networks feature modules with different receptive fields and resolutions, designed to extract diverse representations including different amounts of contextual cues. 
While for known objects the different cues mostly align with the model's notion of a semantic class, in the case of novel objects the representations at multiple context levels tend to disagree.
This can be used as an indicator for uncertainty.
Indeed, our approach develops on this idea by having multiple heads as probes for comprehensive multi-scale cues, and obtains an aggregated uncertainty estimate for out-of-distribution (OoD) segmentation. We show that this strategy improves uncertainty estimates over using a single global prediction and often even over regular ensembles, while being substantially more efficient than the latter. It also sets up the bar on the common benchmarks for OoD segmentation. 
We call our model \modelname{}, for \textbf{M}ulti-head \textbf{Oo}D \textbf{Se}gmentation. Source code available at \url{https://github.com/MOoSe-ECCV22/moose_eccv2022}.

%% file: figures/drawings/teaser.tex
\begin{figure}
    \centering
    \includegraphics[width=0.9\textwidth]{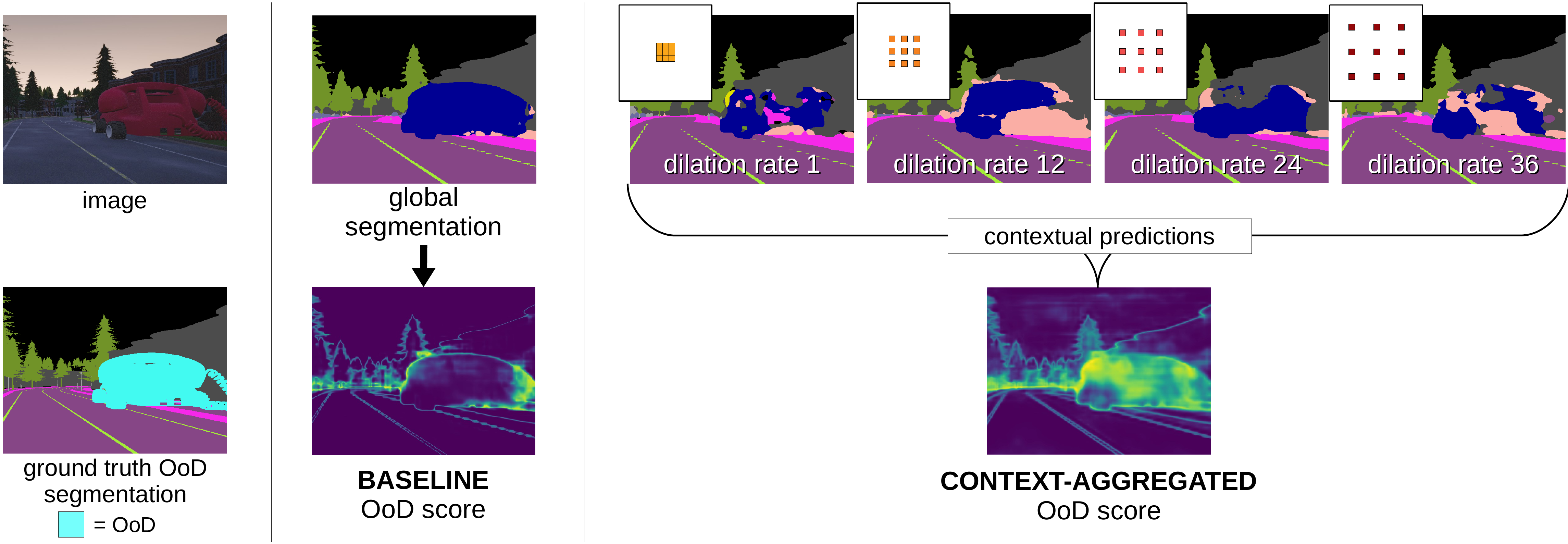}
    \caption{
    Our method obtains predictions based on diverse contextual information from different dilated convolutions, exploiting the hierarchical structure of semantic segmentation architectures. On the anomalous pixels ({\color{cyan} cyan} in the ground truth) the contextual predictions diverge, allowing us to improve upon the global model's uncertainty score, which is overconfident in classifying the object as a car.
    From improved uncertainty we get better out-of-distribution detection. More details in Figure~\ref{fig:architecture} and Section~\ref{subsec:probing}.
    }
    \label{fig:teaser}
\end{figure}

%% file: sections/2_related_work.tex
\subsubsection{Out of Distribution Detection}

Out-of-distribution detection is closely related to uncertainty estimation. 
Under the assumption that a model should be uncertain about samples far from its training distribution, model uncertainty can be used as a proxy score for detecting outliers~\cite{hendrycks17baseline,deep_ensembles}. Several methods for OoD detection, including ours, rely on an existing model trained on a semantic task on the in-distribution data, such as image classification or segmentation~\cite{hendrycks17baseline,a_benchmark}.
Different techniques have been developed to improve outlier detection by means of uncertainty scores, either at inference time~\cite{liang2018enhancing,a_benchmark} or while learning the representations~\cite{hein2019relu,hendrycks2019selfsupervised,bergman2019classification,winkens2020contrastive,hendrycks2018deep}.

Another set of methods for uncertainty estimation is inspired by Bayesian neural networks, which produce probabilistic predictions~\cite{blundell2015weight}.
For example Monte-Carlo Dropout~\cite{gal2016dropout,kendall2017bayesian} approximates the predictive distribution by sampling a finite number of parameter configurations at inference time using random dropout masks.
Although arguably not strictly Bayesian, ensembles~\cite{deep_ensembles,Vyas_2018_ECCV} also approximate the predictive distribution by polling a set of independently trained models fixed at inference time.
Attributes of the predictive distribution, such as its entropy, can be used as a measure for uncertainty~\cite{smith2018understanding,abdar2020review}. 

Alternatively, one can directly model the distribution of the training data, and the likelihood estimated by the resulting model can be used to detect outliers~\cite{nalisnick2018deep,openhybrid,schirrmeister2020understanding,kirichenko2020normalizing}.
Other approaches rather rely on learning pretext tasks on the in-distribution data as a proxy for density estimation. Examples of such tasks are reconstruction~\cite{nguyen2019anomaly,lis2019detecting,xia2020synthesize,kong2021opengan,di2021pixel} and classification of geometric image transformations~\cite{golan2018deep}. 

\subsubsection{Dense OoD Detection}
Methods that are effective at recognizing outlier images do not always scale well to dense OoD detection, where individual pixels in each image need to be classified as in-distribution or anomalous.
A recent work~\cite{a_benchmark} has found that advanced methods like generative models~\cite{baur2018deep,haselmann2018anomaly,schlegl2017unsupervised} and Monte-Carlo dropout~\cite{gal2016dropout} are outperformed by metrics derived from the predictions of a pre-trained semantic segmentation network, such as the values of the segmentation logits. 
Several recent works~\cite{grcic2020dense,meta-ood,bevandic2019simultaneous,cen2021deep} focus on the improvement of such segmentation by-products. In particular, the practice of Outlier Exposure~\cite{hendrycks2018deep}, originally developed for recognition, has recently gained popularity in dense anomaly detection: several approaches revolve around using outlier data during training, either from a real data~\cite{meta-ood,bevandic2019simultaneous,vojir2021road} or sampled from a generative model~\cite{grcic2020dense,kong2021opengan}. While our method does not need outlier exposure to work, we show that it can be beneficially combined with it.

As mentioned above, deep ensembles~\cite{deep_ensembles} are a versatile tool and a gold standard for uncertainty estimation, making them a popular choice for anomaly detection~\cite{Vyas_2018_ECCV,lee2018simple}. Their relative scalability and effectiveness made them a viable option for uncertainty estimation in dense contexts, including anomaly segmentation~\cite{OVNNI,franchi2020tradi}: ensembles are a simple and almost infallible way of improving the quality of uncertainty scores of neural networks.


At the core of ensemble techniques is diversity between models, which is often provided by random weight initialization and data bootstrapping \cite{deep_ensembles,ilg2018uncertainty}, sometimes by architectural differences~\cite{zaidi20}.
While these sources of diversity are of proven efficacy and versatility, they are generic and ignore the requirements of the task at hand, introducing significant computational costs.
Multi-headed ensembles mitigate this drawback by sharing the largest part of the network and drawing their diversity only from independent, lightweight heads~\cite{lee2015m,li2019ensemblenet,Narayanan2021b}.
Even though it is related to multi-headed ensembles, our method exploits an additional source of diversity: leveraging a widespread architectural design specific to semantic segmentation, \modelname{} captures the variety of contextual information and receptive field within the same model. 
This allows for performance improvements that are equal or superior to those of bootstrapped ensembles, at a fraction of the computational cost.

%% file: sections/3_method.tex
\input{figures/drawings/architecture}


\subsection{Contextual Diversity in Semantic Segmentation Networks}
Consider a semantic segmentation decoder based on the popular spatial pyramid architecture~\cite{zhao2017pyramid,chen2017deeplab,chen2017rethinking,cheng2020panoptic}, i.e. containing either a Spatial Pyramid Pooling (SPP~\cite{yoo2015multi}) or Atrous Spatial Pyramid Pooling (ASPP~\cite{chen2017deeplab}) structure. Such structures include a series of pooling or dilated convolutional operations applied in parallel to a set of feature embeddings. We refer to these operations as \textit{spatial pyramid modules}.
Each spatial pyramid module has a unique pooling scale or receptive field, allowing for scale invariance and providing the decoder with various degrees of context: the larger the receptive field the more global context at the cost of a loss in detail.
The outputs of the spatial pyramid modules are concatenated and fed to a segmentation head that produces a single prediction.

In this section we propose a method for extracting the contextual diversity between the representations produced by the different spatial pyramid modules. By exploiting said diversity we are able to improve OoD segmentation performance and network calibration without the need of expensive ensembling of multiple models.
Our method is non-invasive with regard to the main semantic segmentation task and lightweight in terms of computational cost, making it suitable for real-world applications without affecting segmentation performance. In addition, since our approach improves the uncertainty scores of an existing model, it can be easily combined with other state of the art approaches.

\subsection{Probing Contextual Diversity}
\label{subsec:probing}
We start from a generic semantic segmentation architecture featuring $K$ \textit{spatial pyramid modules} $\mathcal{F} = \{f_{c_1}, ..., f_{c_K}\}$, each with a different context size $c_k$, as in Figure~\ref{fig:architecture}.
The encoder features $\boldsymbol\theta$ are fed to the pyramid modules producing a set of contextual embeddings $\Phi = \{\boldsymbol\phi_{c_1}, ..., \boldsymbol\phi_{c_K}\}$, where $\boldsymbol\phi_{c_k} = f_{c_k}(\boldsymbol\theta)$.
The segmentation output is produced by the \textit{global head} $h_g$, which takes all the concatenated features $\Phi$. We denote the output logits of the global head as $\mathbf{z}_g := h_g(\Phi)$.

Our method consists of a simple addition to this generic architecture, namely the introduction of \textit{probes} to extract context-dependent information from the main model. We pair each spatial pyramid module $f_{c_k}$ with a contextual prediction head $h_{c_k}$, which is trained to produce segmentation logits $\mathbf{z}_{c_k} = h_{c_k}(\boldsymbol\phi_{c_k})$ from the context features of size $c_k$ (and the global pooling features, see~\cite{chen2017rethinking}).
Given the input image $\mathbf{x}$, we denote the prediction distributions as:
\begin{equation}
    p(\hat{\mathbf{y}}|\mathbf{x},h_g)=\mathrm{softmax}(\mathbf{z}_h) \hspace{17px} \text{and} \hspace{17px} p(\hat{\mathbf{y}}|\mathbf{x},h_{c_k})=\mathrm{softmax}(\mathbf{z}_{c_k})
\end{equation}
for the global and contextual heads respectively.

We train all the heads using a standard Cross Entropy loss given $N$ classes and the ground truth segmentation $\mathbf{y}$ (we drop the mean operation over the spatial dimensions for simplicity):
\begin{equation}
    \mathcal{L}_\mathrm{CE}(\mathbf{x},\mathbf{y}) = -\sum_{k=1}^{K} \sum_{i=1}^{N} \mathbf{y}_i\cdot\log p(\hat{\mathbf{y}}_i|\mathbf{x},h_{c_k}),
\end{equation}
although any semantic segmentation objective could be used.

The contextual heads are designed to act as probes and extract information from the context-specific representations. For this reason, we do not back-propagate the gradients coming from the contextual heads to the rest of the network, and only update the weights of the heads themselves.
The spatial pyramid modules are distinct operations, each with its specific scope depending on its context size. By stopping the gradients before the spatial pyramid modules we force each head to solve the same segmentation task but using different features, preserving prediction diversity.
As a byproduct, our architectural modifications do not interfere with the rest of the network and the main segmentation task.

\subsubsection{Head Architecture}
The architecture of the contextual heads is based on the global head of the base segmentation model.
For example, for DeepLabV3 it consists of a projection block to bring the number of channels down to 256, followed by a sequence of $d$ prediction blocks ($3\times3$ convolution, batch normalization, ReLU), plus a final $1\times1$ convolution for prediction. The head depth $d$ can be tuned according to the predictive power necessary to process contextual information, depending on the difficulty of the dataset.

\subsection{Out-of-Distribution Detection with \modelname{}}
\label{subsec:scoring_fn}
A model for dense OoD detection should assign to each location in the input image an anomaly score.
To obtain per-pixel OoD scores we test three scoring functions, applied to the outputs of the segmentation heads: maximum softmax probability (MSP)~\cite{hendrycks17baseline}, prediction entropy (H)~\cite{smith2018understanding} and maximum logit (ML)~\cite{a_benchmark}. We adapt each scoring function to work with predictions from multiple heads.
The maximum softmax probability is computed on the average predicted distribution over all the heads, including the global head:
\begin{equation}
    S_{\rm MSP} = -\max_{i\in[1,N]}\bigg{[}\frac{1}{K+1}\bigg{(}p(\hat{\mathbf{y}}_i|\mathbf{x},h_g)+\sum_{k=1}^{K} p(\hat{\mathbf{y}}_i|\mathbf{x},h_{c_k})\bigg{)}\bigg{]},
\end{equation}
Similarly, for the entropy we compute the entropy of the expected output distribution:
\begin{equation}
    S_{\rm H} = \mathcal{H}\bigg{[}\frac{1}{K+1}\bigg{(}p(\hat{\mathbf{y}}|\mathbf{x},h_g)+\sum_{k=1}^{K} p(\hat{\mathbf{y}}|\mathbf{x},h_{c_k})\bigg{)}\bigg{]},
\end{equation}
where $\mathcal{H}$ denotes the information entropy.
For maximum logit we average the logits over the different heads and compute their negated maximum:
\begin{equation}
    S_{\rm ML} = -\max_{i\in[1,N]}\bigg{[}{\frac{1}{K+1}\bigg{(}\mathbf{z}_{g,i}+\sum_{k=1}^{K} \mathbf{z}_{c_k,i}\bigg{)}}\bigg{]}.
\end{equation}
All scores should be directly proportional to the model's belief of a pixel belonging to an anomalous object, therefore for MSP and ML the negatives are taken.

%% file: figures/drawings/architecture.tex
\begin{figure*}
     \centering
     \begin{subfigure}[b]{0.42\textwidth}
         \centering
         \includegraphics[width=\textwidth]{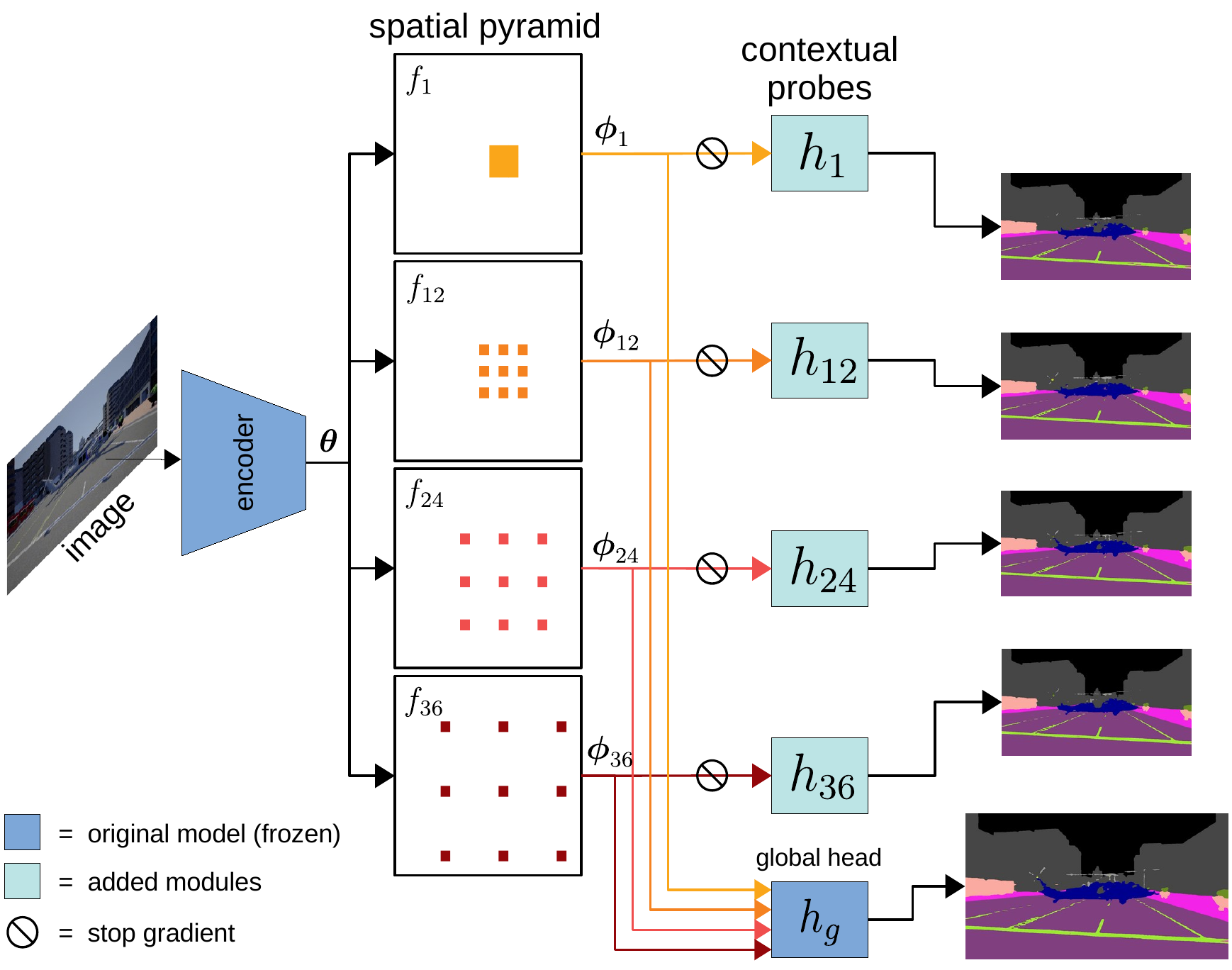}
         \caption{Training pipeline}
     \end{subfigure}
    \hspace{30pt}
     \begin{subfigure}[b]{0.42\textwidth}
         \centering
         \includegraphics[width=\textwidth]{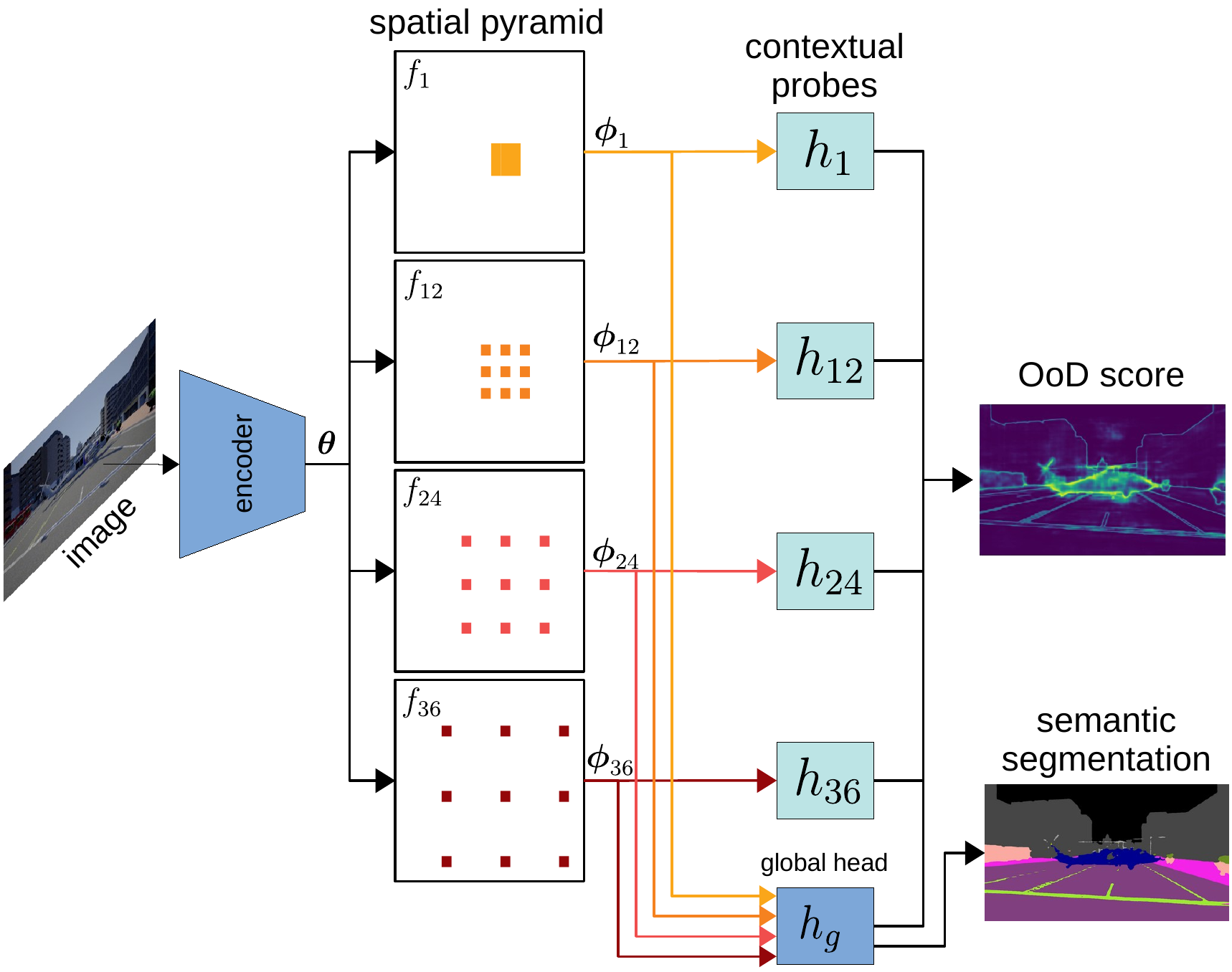}
         \caption{Evaluation pipeline}
     \end{subfigure}
 \caption{\textbf{\modelname}: illustration of the multi-head architecture based on the DeepLabV3 semantic segmentation model.
 During training (a) all the probes learn standard semantic segmentation, while the rest of the network (pre-trained) is unaffected. At test time (b) the uncertainties of all heads (contextual and global) are pooled together into an improved scoremap for OoD detection.
}

 \label{fig:architecture}
\end{figure*}

%% file: sections/4_main_results.tex
\label{sec:experiments}

In this section we evaluate our approach on out-of-distribution detection, comparing it to ensembles (Section~\ref{subsec:exps_ensembles}) and to the state of the art (Section~\ref{subsec:exps_sota}).

\subsection{Datasets \& Benchmarks}\label{datasets}

\textbf{StreetHazards}~\cite{a_benchmark} is a synthetic dataset for semantic segmentation and OoD detection. It features street scenes in diverse settings, created with the CARLA simulation environment~\cite{dosovitskiy2017carla}. 
The 1500 test samples feature instances from 250 different anomalous objects, diverse in appearance, location, and size.

The \textbf{BDD-Anomaly}~\cite{a_benchmark} dataset is derived from the BDD100K~\cite{yu2018bdd100k} semantic segmentation dataset by removing the samples containing instances of the motorcycle, bicycle, and train classes, and using them as a test set for OoD segmentation, yielding a 6280/910/810 training/validation/test split. BDD-Anomaly and StreetHazards constitute the CAOS benchmark~\cite{a_benchmark}.

\textbf{Fishyscapes - LostAndFound}~\cite{pinggera2016lost,blum2019fishyscapes} is a dataset for road obstacle detection, designed to be used in combination with the Cityscapes~\cite{Cordts2016Cityscapes} driving dataset.
Its test split contains 1203 images of real street scenes featuring road obstacles, whose presence is marked in the segmentation ground truth. 

\textbf{RoadAnomaly}~\cite{lis2019detecting} consists of 60 real world images of diverse anomalous objects in driving environments, collected from the internet. The images come with pixel-wise annotations of the anomalous objects, making them suitable for testing models trained on driving datasets.

Results for the anomaly track of the  {SegmentMeIfYouCan}~\cite{segmentmeifyoucan2021} benchmark can be found in the supplementary material.

\subsection{Evaluation Metrics}
We evaluate OoD detection performance using the area under the precision-recall curve (AUPR), and the false positive rate at 95\% true positive rate (FPR$_{95}$). As is customary, anomalous pixels are considered positives.
Other works include results for the area under the ROC curve (AUROC), however the AUPR is to be preferred to this metric in the presence of heavy class imbalance, which is the case for anomaly segmentation~\cite{davis2006relationship}.

\input{tables/ensembles_fn_first_psp}

\subsection{Experimental Setup}\label{subsec:exp_setup}
\modelname\ relies on semantic segmentation models to perform dense OoD detection. For the experiments on StreetHazards and BDD-Anomaly we report results for two convolutional architectures, (DeepLabV3~\cite{chen2017deeplab} and PSPNet~\cite{zhao2017pyramid}, each with ResNet50 and ResNet101 backbones) and one transformer based (Lawin~\cite{Lawin}, see supplementary material).
For the experiments on LostAndFound and RoadAnomaly we use DeepLabV3+~\cite{deeplabv3plus2018} with a ResNet101 backbone, trained on Cityscapes\footnote{Parameters available at:\\ \url{https://github.com/NVIDIA/semantic-segmentation}} or BDD100k~\cite{yu2018bdd100k} respectively.

\subsubsection{Training}
We build \modelname{} on top of fully trained semantic segmentation networks, by adding the prediction heads and training them jointly for segmentation on the respective dataset, using a standard pixel-wise cross-entropy loss. Although nothing prevents from training the whole model together, for fairness of comparison we only apply the loss to the probes.
In order to prevent any alteration to the main model while training the heads, we stop gradient propagation through the rest of the network and make sure that the normalization layers would not update their statistics during forward propagation. The heads are trained for 80 epochs, or until saturation of segmentation performance (mIoU).

\modelname{} introduces two hyperparameters: learning rate and depth $d$ of the contextual heads. By default we use $d=1$ for the models trained on StreetHazards and $d=3$ otherwise.
While the performance gains depend on these, we find that our method is robust to configuration changes, as we show in an ablation study in the supplementary material.

\subsection{Comparison with Ensembles}
\input{figures/samples/qualitative_examples_SH_2}

\label{subsec:exps_ensembles}
In this section we compare \modelname{} with the single prediction baseline (global head) and with two types of ensembles. Deep ensembles~\cite{deep_ensembles} (\deepens) consist of sets of independent segmentation networks, each trained on a different random subset of $67\%$ of the original data, starting from a different random parameter initialization~\cite{ilg2018uncertainty}. Similarly, multi-head ensembles (\multiheadens) are trained on random data subsets, but share the same encoder and only feature diverse prediction heads, for increased efficiency.

We compare to ensembles with 5 members/heads to match the number of heads in our method. Additionally, we pick the ensemble member with the median AUPR performance to serve both as the single model baseline and as initialization for \modelname{}. The shared backbone of the multi-head ensembles also comes from the same model.

Table~\ref{tab:ensembles} shows results for the CAOS benchmark (StreetHazards and BDD-Anomaly) using DeepLabV3 and PSPNet as base architectures, with ResNet50~\cite{he2016deep} backbones. We report results for the three OoD scoring functions described in Section~\ref{subsec:scoring_fn}; results for \modelname{} are averaged over 3 runs, standard deviations are available in the supplementary material. For all datasets, architectures, scoring functions, and metrics, \modelname{} consistently outperforms its respective global head. Similarly, \modelname{} outperforms multi-head ensembles, as well as deep ensemble in most cases, while having a smaller computational cost than both.

In accordance with what observed in other works~\cite{a_benchmark}, the maximum-logit scoring function tends to outperform entropy, most notably in terms of FPR$_{95}$ and on the BDD-Anomaly dataset. Both scoring functions consistently outperform maximum-softmax probability. Moreover, maximum-logit appears to combine well with \modelname{} by effectively reducing false positives.
Results for models using the ResNet101 backbone are available in the supplementary material.

Figure \ref{fig:qualitative_samples} shows an example for OoD segmentation on a driving scene. The top row compares the entropy obtained using the global head (\ref{fig:5_esg}) and our multi-head approach (\ref{fig:5_pred}).
The probes of \modelname{} disagree on the nature of the anomalous objects in the image, and its aggregated entropy score is able to outline the anomalous objects more clearly than the global head. However, prediction disagreement also produces false positives for smaller inlier objects, such as the street sign on the right, highlighting a possible failure mode of our approach.

\subsubsection{Computational costs}
\input{tables/runtimes}
In Table~\ref{tab:runtimes} we compare our method against ensembles in terms of computational costs, reporting the number of parameters of each model and the estimated runtime of a forward pass. We consider DeepLabV3 and PSPNet with ResNet50 and \modelname{} head depth 1.
Deep ensembles have the highest parameter count and runtime, 5 times that of a single network. \modelname{} compares favorably to both ensembles on all architectures. The larger size and runtime of PSPNet compared to DeepLab is due to its higher dimensional representations, which can be reduced with projection layers before the probes.

\subsection{Comparison with the State of The Art}
Here we compare \modelname{} with the best approaches for dense OoD detection that do not require negative training data (see Section \ref{sec:OE}).
\label{subsec:exps_sota}
\subsubsection{The CAOS Benchmark}
On StreetHazards and BDD-Anomaly we compare with TRADI~\cite{franchi2020tradi}, SynthCP~\cite{xia2020synthesize}, OVNNI~\cite{OVNNI}, Deep Metric Learning (DML)~\cite{cen2021deep}, and the approach by Grcic et al.~\cite{grcic2020dense} that uses outlier exposure with generated samples.
TRADI and OVNNI require multiple forward passes per sample, increasing the evaluation run-time (or memory requirements) considerably.
Table~\ref{tab:SOTA}(a) shows that \modelname{} compares favorably to existing works on both datasets and on all metrics. We note that, given its non-invasive nature, \modelname{} is compatible with other approaches, and can for example be combined with the loss of DML.

\input{tables/SOTA_side_by_side.tex}

\subsubsection{LostAndFound, RoadAnomaly}
We extend our evaluation of \modelname{} to other real world benchmarks, Fishyscapes LostAndFound and RoadAnomaly, using models trained on Cityscapes and BDD100k~\cite{yu2018bdd100k} respectively, as described in Section~\ref{subsec:exp_setup}. We report our results in Table~\ref{tab:SOTA}(b) and include results for the comparable (not needing negative training data) state-of-the-art methods DML, Standardized Max-Logits~\cite{Jung_2021_ICCV} and Image Resynthesis~\cite{lis2019detecting}. Similarly to the CAOS benchmark, we can observe that the adoption of \modelname{} improves OoD detection performance on both benchmarks, regardless of the chosen scoring function. 
In the supplementary material we include results for the SegmentMeIfYouCan benchmark (similar to RoadAnomaly), where Image Resynthesis is currently SOTA. On the other hand, Table~\ref{tab:SOTA}(b) shows that Image Resynthesis does not perform well on LostAndFound.
Standardized Max-Logits has remarkable results on LostAndFound but not on RoadAnomaly, where \modelname{} works best.

Figure \ref{fig:6_failure_cases} shows examples of OoD detection on LostAndFound and RoadAnomaly. 
In the first example \modelname{} improves over the entropy heatmap of the global head. In the second example, however, it can be seen that our method still fails to detect the obstacles and produces more false positives. Increased false positives are visible in the background of the third example too, although here \modelname{} also improves the detection of some anomalous objects.

\input{figures/samples/qualitative_examples_LaF}

\subsubsection{Outlier Exposure}
\label{sec:OE}
Several methods for anomaly segmentation rely on negative training data from a separate source. 
While this technique introduces some drawbacks, such as a reliance on the choice of the negative data and a potential negative impact on segmentation, it has been shown to improve OoD detection on the common benchmarks.
Following the procedure described in \cite{meta-ood} as "entropy training", we investigate whether our method can also benefit from outlier exposure. Indeed, results show that outlier exposure boosts MOoSe+ML to 53.19 AUPR (+22\%) and 24.38 FPR$_{95}$ (-24\%) on RoadAnomaly. Full results on all scoring functions are available in the supplementary material.

%% file: tables/ensembles_fn_first_psp.tex
\begin{table*}[ht]
\caption{\textbf{CAOS benchmark.} Comparison between single model/global head (\globalhead), multi-head ensembles (\multiheadens), standard deep ensembles (\deepens) and \modelname\ on dense out-of-distribution detection. 
The results are for DeepLabV3 and PSPNet models with ResNet50 backbones. All three scoring functions (maximum softmax probability (MSP), entropy (H), maximum logit (ML)) are considered. All results are percentages, best results are shown in \textbf{bold}}
\centering
\scriptsize
\begin{tabular}{ll|rrrr|rrrr}
\toprule
    &  & \multicolumn{4}{c|}{StreetHazards}    & \multicolumn{4}{c}{BDD-Anomaly}\\
    &  & \multicolumn{2}{c}{DeepLabV3}     & \multicolumn{2}{c|}{PSPNet}  & \multicolumn{2}{c}{DeepLabV3}     & \multicolumn{2}{c}{PSPNet}  \\
    \multicolumn{1}{c}{Score fn.}& Method & \multicolumn{1}{c}{AUPR$\uparrow$} & \multicolumn{1}{c}{FPR$_{95}\downarrow$} & \multicolumn{1}{c}{AUPR$\uparrow$} & \multicolumn{1}{c|}{FPR$_{95}\downarrow$} & \multicolumn{1}{c}{AUPR$\uparrow$} & \multicolumn{1}{c}{FPR$_{95}\downarrow$} & \multicolumn{1}{c}{AUPR$\uparrow$} & \multicolumn{1}{c}{FPR$_{95}\downarrow$} \\
\midrule
MSP & \globalhead   & 9.11    & 22.37  & 9.65  & 22.04   & 7.01    & 22.47   & 6.75     & 23.63  \\
    & \multiheadens & 9.69    & 21.40  & 9.84  & 22.49   & 7.55    & 25.50   & 8.07     & 23.41  \\
    & \deepens      & 10.22   & 21.09  & 10.61 & \textbf{20.75}   & 7.64    & \textbf{21.53}   & \textbf{8.52}     & \textbf{21.31}  \\
    & \modelname    & \textbf{12.53}   & \textbf{21.05}  & \textbf{11.28} & 21.94   & \textbf{8.66}    & 22.49   & 8.11	    & 24.09  \\
\midrule
H   & \globalhead       & 11.89   & 22.07  & 12.28   & 21.77   & 10.23   & 20.64  & 9.89    & 21.69  \\
    & \multiheadens     & 12.59   & 21.10  & 12.45   & 22.29  & 10.62   & 23.51  & 11.73   & 20.76  \\
    & \deepens          & 13.43   & 20.62  & 13.39   & \textbf{20.35}   & 11.39   & 19.31  & 12.32   & \textbf{18.83}  \\
    & \modelname        & \textbf{15.43}   & \textbf{19.89}  & \textbf{14.52}	 & 21.20   & \textbf{12.59}   & \textbf{19.27}  & \textbf{12.35}   & 20.98  \\
\midrule
ML  & \globalhead   & 13.57   & 23.27  & 13.43   & 27.71   & 10.69   & 15.60  & 10.68   & 16.79  \\
    & \multiheadens & 13.99   & 21.86  & 13.64	 & 28.30   & 10.69   & 20.19  & 12.40   & 15.08  \\
    & \deepens      & 14.57   & 21.79  & 14.14   & 25.82   & 11.40   & 14.66  & 12.26   & 13.96  \\
    & \modelname    & \textbf{15.22}   & \textbf{17.55}  & \textbf{15.29}   & \textbf{20.46}   & \textbf{12.52}   & \textbf{13.86}  & \textbf{12.88}   & \textbf{13.94} \\

\bottomrule
\end{tabular}
\label{tab:ensembles}
\end{table*}

%% file: figures/samples/qualitative_examples_SH_2.tex
\begin{figure*}
     \centering
     \begin{subfigure}[b]{0.20\textwidth}
         \centering
         \includegraphics[width=\textwidth]{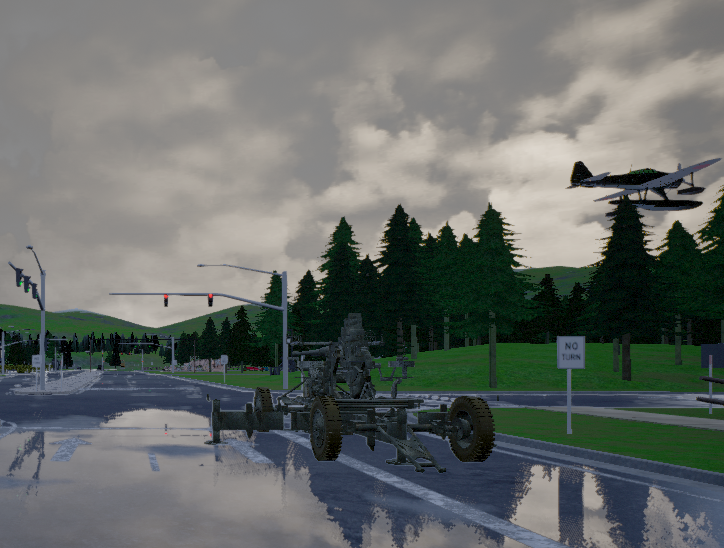}
         \caption{\scriptsize Image}
         \label{fig:5_img}
     \end{subfigure}
     \hfill
     \begin{subfigure}[b]{0.20\textwidth}
         \centering
         \includegraphics[width=\textwidth]{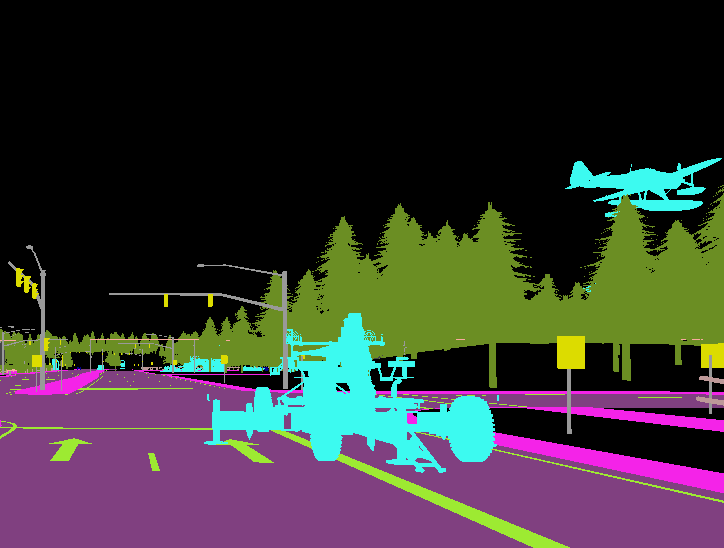}
         \caption{\scriptsize Ground truth}
         \label{fig:5_gt}
     \end{subfigure}
     \hfill
     \begin{subfigure}[b]{0.20\textwidth}
         \centering
         \includegraphics[width=\textwidth]{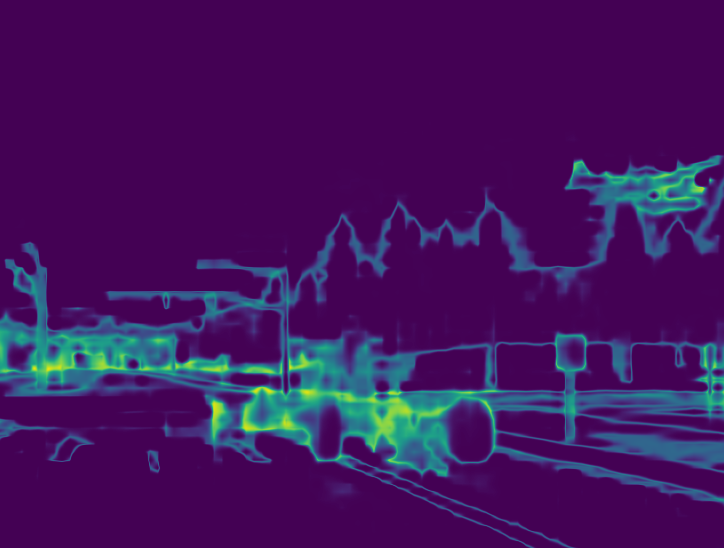}
         \caption{\scriptsize Entropy $h_g$}
         \label{fig:5_esg}
     \end{subfigure}
     \hfill
     \begin{subfigure}[b]{0.20\textwidth}
         \centering
         \includegraphics[width=\textwidth]{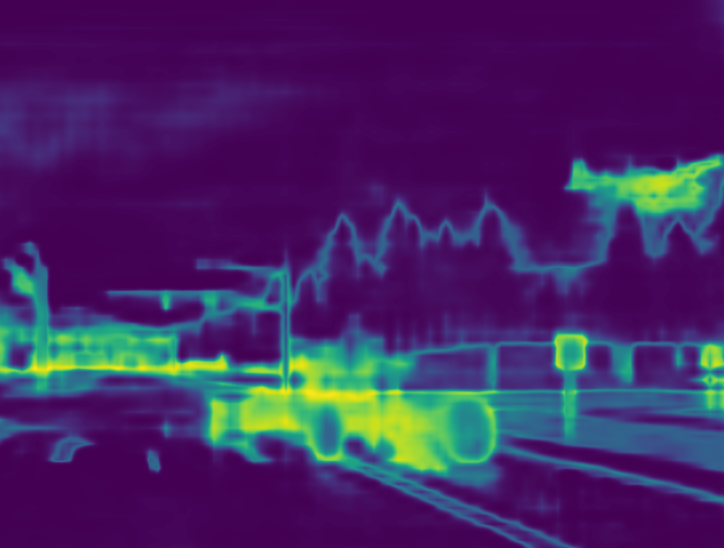}
         \caption{\scriptsize Ent. \modelname}
         \label{fig:5_pred}
     \end{subfigure}
    
     \begin{subfigure}[b]{0.16\textwidth}
         \centering
         \includegraphics[width=\textwidth]{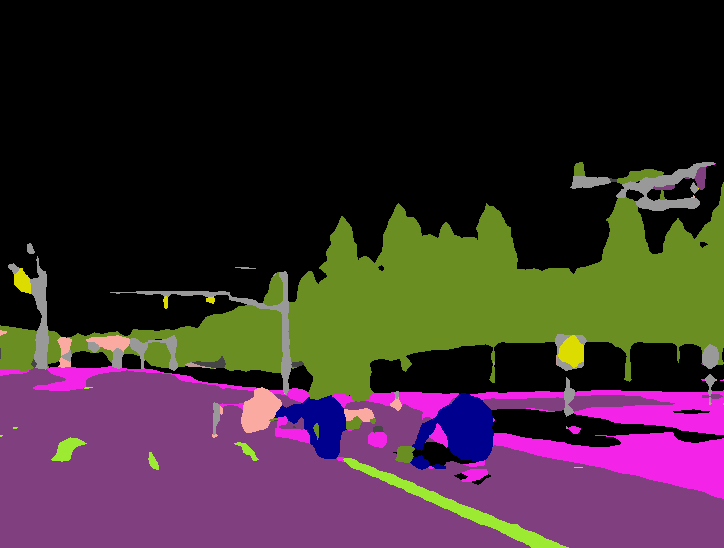}
         \caption{\scriptsize Segm. $s_1$}
         \label{fig:5_s1}
     \end{subfigure}
     \hfill
     \begin{subfigure}[b]{0.16\textwidth}
         \centering
         \includegraphics[width=\textwidth]{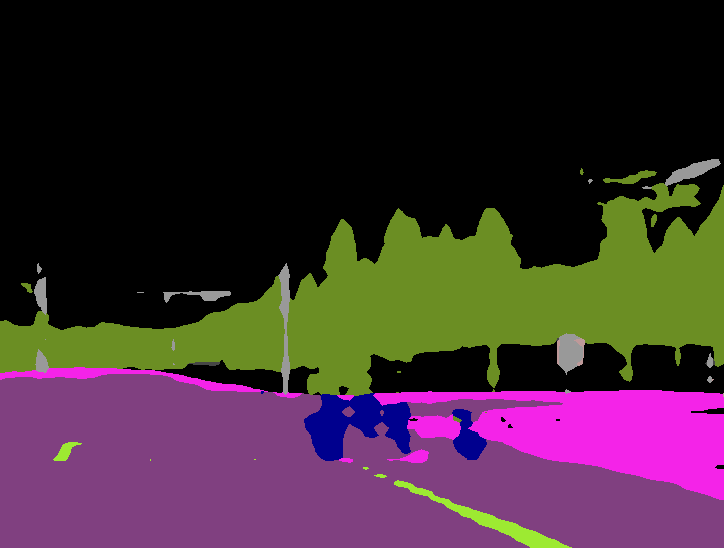}
         \caption{\scriptsize Segm. $s_{12}$}
         \label{fig:5_s12}
     \end{subfigure}
     \hfill
     \begin{subfigure}[b]{0.16\textwidth}
         \centering
         \includegraphics[width=\textwidth]{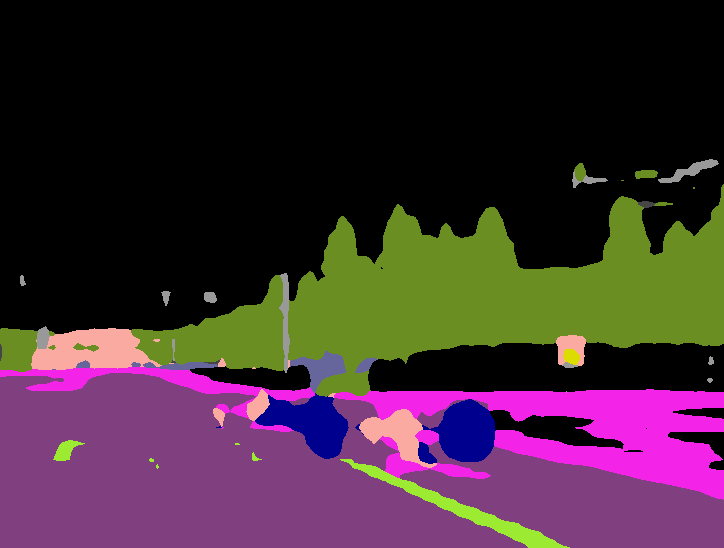}
         \caption{\scriptsize Segm. $s_{24}$}
         \label{fig:5_s24}
     \end{subfigure}
     \hfill
     \begin{subfigure}[b]{0.16\textwidth}
         \centering
         \includegraphics[width=\textwidth]{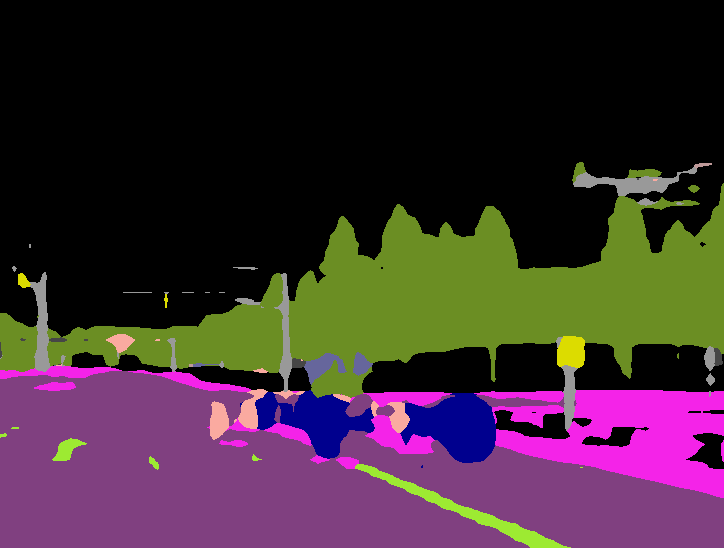}
         \caption{\scriptsize Segm. $s_{36}$}
         \label{fig:5_s36}
     \end{subfigure}
     \hfill
     \unskip\ \vrule\
     \hfill
     \begin{subfigure}[b]{0.16\textwidth}
         \centering
         \includegraphics[width=\textwidth]{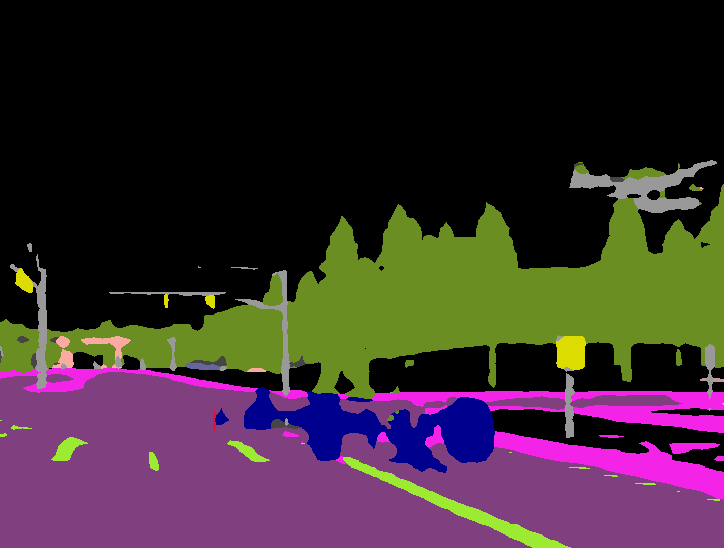}
         \caption{\scriptsize Segm. $s_g$}
         \label{fig:5_sg}
     \end{subfigure}
    \caption{\textbf{OoD segmentation \& head contribution.} 
    Test image from the StreetHazards dataset: a street scene containing anomalous objects (indicated in {\color{cyan} cyan} in the ground truth (b)). The contextual predictions (e-h) diverge on the outliers, improving the entropy score map (c, d). The example shows an interesting failure case: the street sign (in-distribution) on the right also sparks disagreement between the heads, resulting in increased entropy and thus a false positive.
    }
    \label{fig:qualitative_samples}
\end{figure*}


%% file: tables/runtimes.tex
\begin{table}
\tiny
\caption{\textbf{Computational costs.} Estimated computational costs of \modelname{} in comparison with ensembles. We report the number of parameters (in millions) and the estimated forward pass runtime on StreetHazards (in milliseconds), estimated on a single Nvidia RTX2080Ti GPU using the PyTorch~\cite{paszke2017automatic} benchmarking utilities}
\centering
\begin{tabular}{llrrrr}
\toprule
          Architecture & & \multicolumn{1}{@{\hspace{10px}}c}{Single}       & \multicolumn{1}{@{\hspace{10px}}c}{\multiheadens}       & \multicolumn{1}{@{\hspace{10px}}c}{\deepens}        & \multicolumn{1}{@{\hspace{10px}}c}{MOoSe} \\
\midrule
DeepLabV3 & parameters (M)           & 40       & 104      & 198       & 43     \\
          & runtime (ms)             & 113      & 286      & 583       & 121    \\
PSPNet    & parameters (M)           & 47       & 139      & 233       & 94      \\
          & runtime (ms)             & 107      & 246      & 542       & 183     \\
\bottomrule         
\end{tabular}
\label{tab:runtimes}
\end{table}

%% file: tables/SOTA_side_by_side.tex
\begin{table}
\caption{\textbf{State-of-the-art comparison.} \textbf{Left - CAOS benchmark:} \modelname{} in combination with the max-logit scoring function, outperforms all other methods on StreetHazards, except for DML in terms of FPR$_{95}$. On BDD-Anomaly \modelname{} performs the best in both metrics. \textbf{Right}: \modelname{} yields improvements on both \textbf{Fishyscapes LostAndFound (FS - LaF)} and \textbf{RoadAnomaly}, but on the former benchmark is outperformed by Standardized Max-Logits (Std.ML).}
\setlength{\tabcolsep}{2pt}
\scriptsize
\hspace{10px}
\begin{minipage}{.4\linewidth}
{
\renewcommand{\arraystretch}{1.12}
\setlength{\tabcolsep}{1pt}
\begin{tabular}{l|rr|rr}
\toprule
            
        
             & \multicolumn{2}{c|}{\textbf{Street}}   & \multicolumn{2}{c}{\textbf{BDD}}     \\
             & \multicolumn{2}{c|}{\textbf{Hazards}}   & \multicolumn{2}{c}{\textbf{Anomaly}}     \\
             \rule{0pt}{14px}
             & AUPR & FPR$_{95}$ & AUPR & FPR$_{95}$ \\
             
             \midrule
TRADI\cite{franchi2020tradi}       & 7.2\phantom{0}      & 25.3\phantom{0}      & 5.6\phantom{0}      & 26.9\phantom{0}     \\
SynthCP\cite{xia2020synthesize}    & 9.3\phantom{0}      & 28.4\phantom{0}      & -\phantom{0}        & -\phantom{0}        \\
OVNNI\cite{OVNNI}                  & 12.6\phantom{0}     & 22.2\phantom{0}      & 6.7\phantom{0}      & 25.0\phantom{0}     \\
Grcic\cite{grcic2020dense}         & 12.7\phantom{0}     & 25.2\phantom{0}      & -\phantom{0}        & -\phantom{0}        \\
DML\cite{cen2021deep}              & 14.7\phantom{0}     & \textbf{17.3}\phantom{0}      & - \phantom{0}       & - \phantom{0}       \\ \midrule
\modelname{} ML                       & \textbf{15.22}    & 17.55     & \textbf{12.52}    & \textbf{13.86}   \\
\bottomrule
\end{tabular}
}
\end{minipage}
\hspace{15px}
\begin{minipage}{.4\linewidth}
{
\setlength{\tabcolsep}{1pt}
\begin{tabular}{ll|rr|rr}
\toprule
 
 & & \multicolumn{2}{@{\hspace{10px}}c@{\hspace{10px}}|}{\textbf{FS - LaF}} & \multicolumn{2}{c}{\textbf{RoadAnomaly}}      \\
 \multicolumn{2}{r|}{Method} & AUPR & FPR$_{95}$ & AUPR & FPR$_{95}$ \\ 
 
 \midrule
MSP & \globalhead           & 3.06	  & 37.46         & 23.76                 & 51.32     \\
& \modelname                & 7.13	  & 33.72         & 31.53                 & 43.41     \\
H & \globalhead             & 6.23    & 37.34         & 32.00                 & 49.14     \\
& \modelname                & 12.08   & 32.58         & \underline{41.48}     & \underline{36.78} \\
ML & \globalhead            & 10.25   & 37.45         & 37.86                 & 39.03     \\
& \modelname                & \underline{13.64}   & \underline{32.32} & \textbf{43.59}        & \textbf{32.12}     \\ 
\midrule
\multicolumn{2}{r|}{Resynth.~\cite{lis2019detecting}} & 5.70 & 48.05 & - & - \\
\multicolumn{2}{r|}{DML~\cite{cen2021deep}} &         - & -             & 37\phantom{.00}   & 37\phantom{.00}    \\ 
\multicolumn{2}{r|}{Std.ML~\cite{Jung_2021_ICCV}}& \textbf{31.05} & \textbf{21.52} & 25.82 & 49.74 \\
\bottomrule
\end{tabular}
}
\end{minipage}
\label{tab:SOTA}
\end{table}

%% file: figures/samples/qualitative_examples_LaF.tex
\begin{figure*}[t]
     \centering
     \begin{subfigure}[b]{0.18\textwidth}
         \centering
         \includegraphics[width=\textwidth]{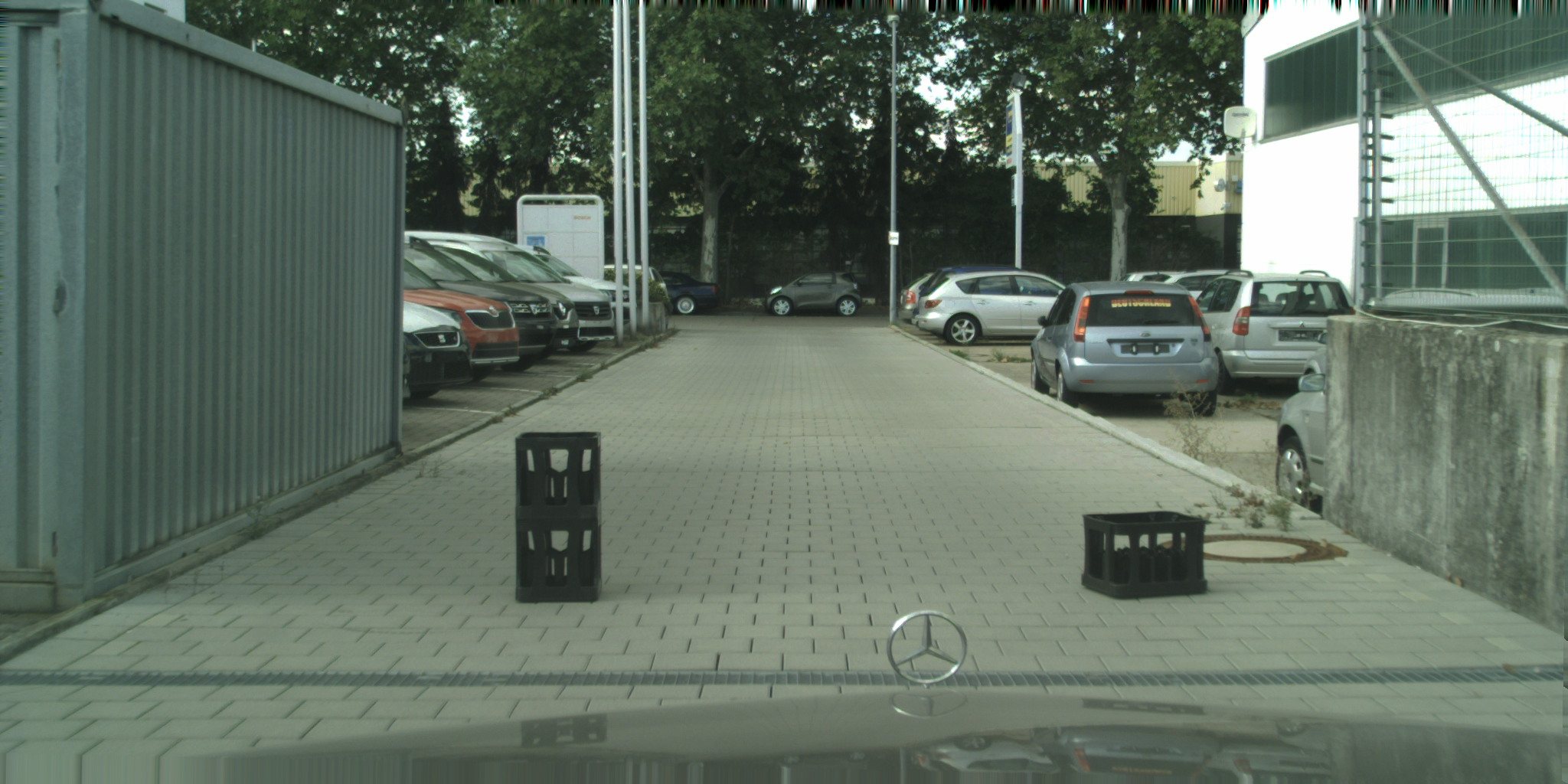}
     \end{subfigure}
     \hfill
     \begin{subfigure}[b]{0.18\textwidth}
         \centering
         \includegraphics[width=\textwidth]{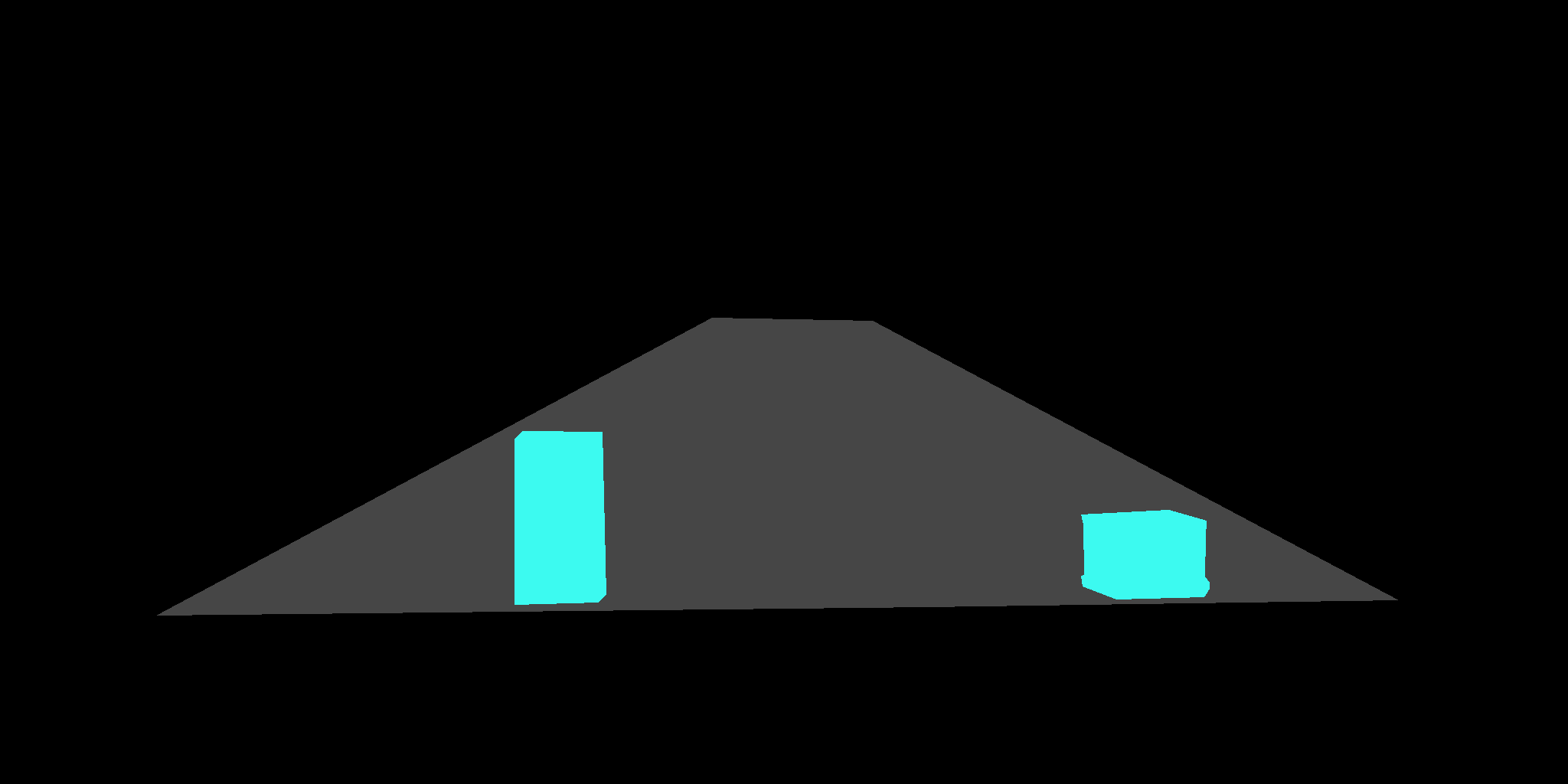}
     \end{subfigure}
     \hfill
     \begin{subfigure}[b]{0.18\textwidth}
         \centering
         \includegraphics[width=\textwidth]{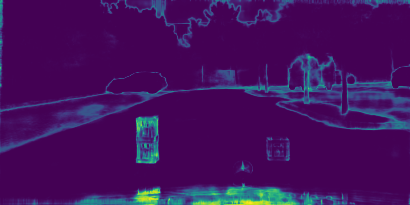}
     \end{subfigure}
     \hfill
     \begin{subfigure}[b]{0.18\textwidth}
         \centering
         \includegraphics[width=\textwidth]{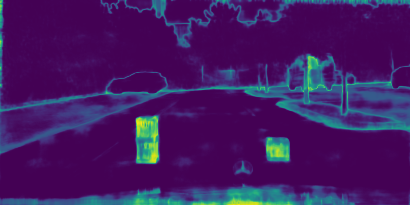}
     \end{subfigure}
     
     \begin{subfigure}[b]{0.18\textwidth}
         \centering
         \includegraphics[width=\textwidth]{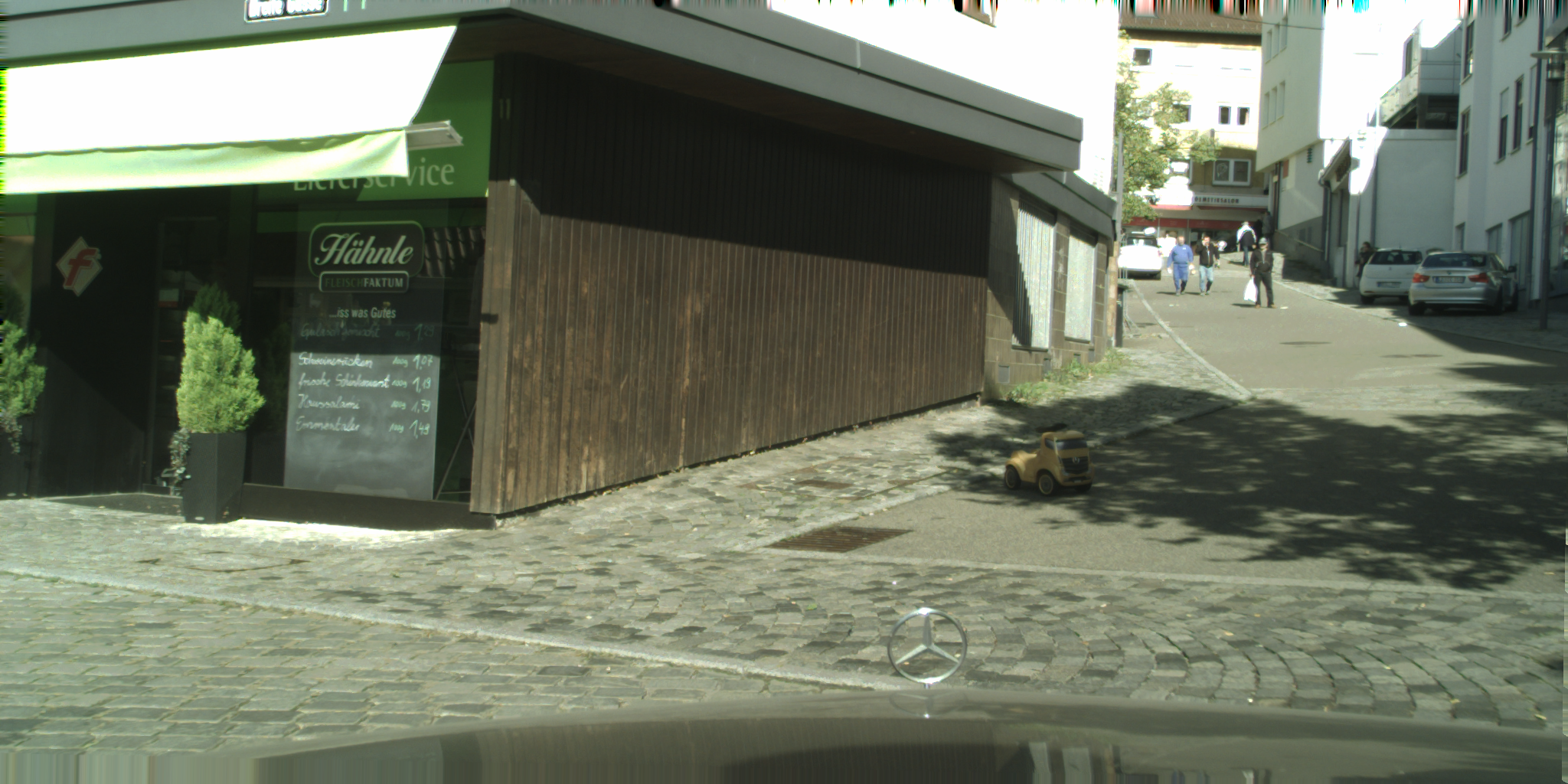}
     \end{subfigure}
     \hfill
     \begin{subfigure}[b]{0.18\textwidth}
         \centering
         \includegraphics[width=\textwidth]{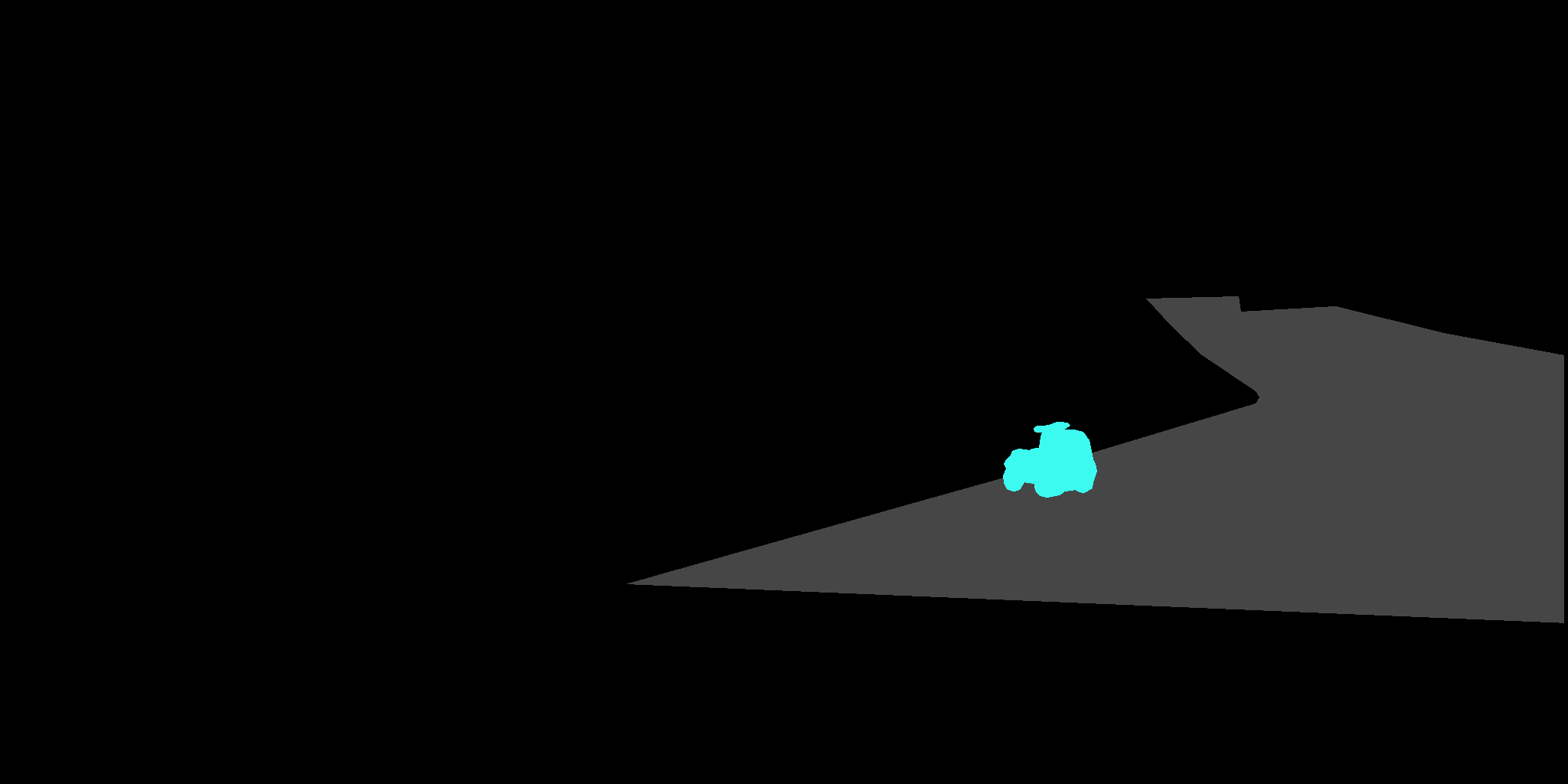}
     \end{subfigure}
     \hfill
     \begin{subfigure}[b]{0.18\textwidth}
         \centering
         \includegraphics[width=\textwidth]{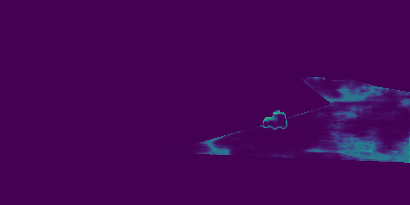}
     \end{subfigure}
     \hfill
     \begin{subfigure}[b]{0.18\textwidth}
         \centering
         \includegraphics[width=\textwidth]{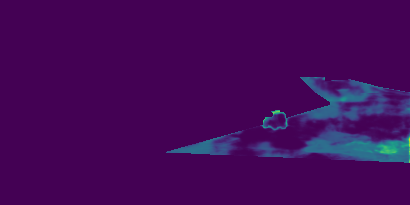}
     \end{subfigure}
     
     \begin{subfigure}[b]{0.18\textwidth}
         \centering
         \includegraphics[width=\textwidth]{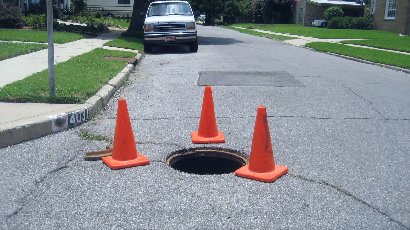}
         \caption{\scriptsize Image}
         \label{fig:ra21_img}
     \end{subfigure}
     \hfill
     \begin{subfigure}[b]{0.18\textwidth}
         \centering
         \includegraphics[width=\textwidth]{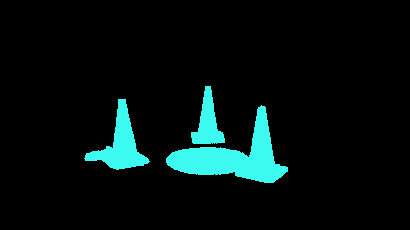}
         \caption{\scriptsize OoD G.T.}
         \label{fig:ra21_gt}
     \end{subfigure}
     \hfill
     \begin{subfigure}[b]{0.18\textwidth}
         \centering
         \includegraphics[width=\textwidth]{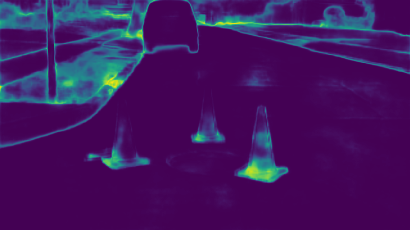}
         \caption{\scriptsize Entropy $h_g$}
         \label{fig:ra21_esg}
     \end{subfigure}
     \hfill
     \begin{subfigure}[b]{0.18\textwidth}
         \centering
         \includegraphics[width=\textwidth]{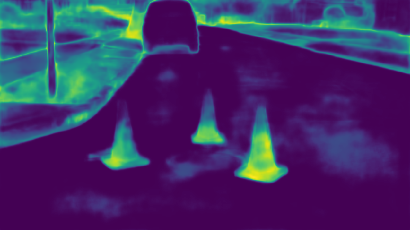}
         \caption{\scriptsize Ent. \modelname}
         \label{fig:ra21_pred}
     \end{subfigure}
 \caption{\textbf{OoD segmentation.} Examples on LostAndFound and RoadAnomaly, anomalous objects shows in cyan in the second column. The first row shows an example in which our model is able to recognize anomalous objects in their entirety, where the global head fails. The second row shows a failure case, where our model only marks the borders of the obstacle on the road and produces false positives, performing worse than the global head. The example in the last row is from RoadAnomaly: \modelname{} detects the traffic cones better than the global head, but introduces noise in the background and still fails to detect the manhole.}
 \label{fig:6_failure_cases}
\end{figure*}

%% file: sections/5_analysis_new.tex
Our approach relies on a collection of different predictions to improve OoD detection. Previous literature on network ensembles puts the spotlight on diversity~\cite{liu2019deep,lee2015m}, emphasizing that multiple estimators can be helpful only if their predictions are diverse and each contributes with useful information for the cumulative decision.
In this section we address some points to better understand the working principle of the method and verify its underlying hypotheses. Specifically, we investigate: 1) the effect of \modelname{} on prediction diversity, 2) whether contextual aggregation can be responsible for prediction diversity, and 3) how this translates into better OoD detection.

\subsection{Quantifying Diversity: Variance and Mutual Information}
\label{sec:variance_mutual_info}
We are interested in comparing \modelname{} to the closely related ensembles in terms of prediction diversity, of which a simple metric is variance. We compare the average variance of the output distributions of \modelname{} and ensembles on StreetHazards and BDD-Anomaly validation, as reported in Table~\ref{tab:ablations} (left). On both datasets our method's predictions have higher variance than both ensembles.

Variance, however, gives us no insights on what the predictions disagree upon, and is therefore of limited interest.
From the literature on Bayesian networks we can borrow a more informative metric: the mutual information (MI) between the model distribution and the output distribution~\cite{malinin2019ensemble}.
Consider an ensemble of $K$ networks, or a multi-head model with $K$ heads. Each model or head produces a prediction $p(\hat{y}|x,k)$.
We can compute the mutual information between the distribution of the models $k$ and the distribution of their predictions as:
\begin{equation}\label{eq:mutual_info}
    \text{MI}(\hat{y},k|x) =  {\mathcal{H}\bigg{[}\frac{1}{K}\sum_{k=1}^{K} p(\hat{y}|x,k)}\bigg{]} - \frac{1}{K}\sum_{k=1}^{K}\mathcal{H}\bigg{[}p(\hat{y}|x,k)\bigg{]},
\end{equation}
which is the entropy of the expected output distribution minus the average entropy of the output distributions. 
MI is high for a sample $x$ if the predictions are \textit{individually confident but also in disagreement with each other}.
This tells us how much additional information the 
diversity brings to the overall model: if all the predictions are equally uncertain about the same samples they disagree on, then aggregating them will not affect the aggregated uncertainty estimate.

In Table~\ref{tab:ablations} (left) we report the average MI on StreetHazards and BDD-Anomaly validation, comparing again \modelname{} and ensembles.
Similarly to variance, our method's predictions have higher MI than both ensemble types, indicating that contextual probing not only produces more diversity in absolute terms, but also that this diversity adds more information to the model's predictive distribution.

Finally, in Table~\ref{tab:ablations} (left) we report the Expected Calibration Error~\cite{guo2017calibration} of all methods, to show that even if ensembles are better calibrated than the baseline, it is \modelname{} that performs the best at uncertainty estimation overall.

\input{tables/ablations_side_by_side}

\subsection{Context as a Source of Diversity}
In the previous section we showed that our approach produces highly diverse predictions. In this section we investigate the source of this diversity: our hypothesis is that each head relies differently on contextual information depending on the dilation rate of their respective spatial pyramid module, resulting in diverse predictive behaviors.

We test this hypothesis by evaluating the ability of each head to perform semantic segmentation when \textit{only} contextual information is available.
We corrupt the pixels of the foreground classes in BDD-Anomaly\footnote{Pole, traffic light, traffic sign, person, car, truck, bus.} with random uniform noise while leaving the background pixels unchanged, then we evaluate how well each head can still classify the corrupted foreground pixels by relying on the context.
An example of the process can be seen in Figure~\ref{fig:fg_corruption} (left).
Figure~\ref{fig:fg_corruption} (right) shows the mIoU on the noisy foreground as a percentage of the foreground mIoU on the original clean image. We can observe that dilation rate and robustness to foreground corruption are proportional to each other at multiple noise levels, as further illustrated by the qualitative example in the figure. The different result quality for different dilation rates confirm the validity of contextual aggregation as a source of prediction diversity, as anticipated by the comparison with regular (non-contextual) ensembles on variance and mutual information in Section~\ref{sec:variance_mutual_info}.

\input{figures/graphs/foreground_noise}

\subsection{Effect of Contextual Diversity on OoD Detection}
The results presented in Section~\ref{sec:experiments} already show that contextual probing improves performance on the task. Moreover, results obtained from the application of \modelname{} to transformer-based models (7.3\% average AUPR increase on StreetHazards across scoring functions), which are available in the supplementary material, indicate that the principle is applicable across architectures and its gains are not an artifact of CNNs.

The last point to address is the contribution of contextual diversity to out-of-distribution detection.
To quantify this contribution, we performed an ablation study removing receptive field diversity from DeepLabV3 by using the same dilation rate for all the convolutions in the spatial pyramid module.
We train several versions of this single-dilation (SD) \modelname{}, each with a different dilation rate, and present the comparison with standard \modelname{} in Table~\ref{tab:ablations} (right).
Firstly, all single dilation models have lower prediction variance than regular \modelname{}.
Secondly, although the single-dilation models still outperform their global head, \modelname{} yields larger gains than all SD models, both in absolute and relative terms.
While these results confirm that contextual diversity is crucial for the success of our method, they also show that there are more contributing factors, compatibly with the known benefits of ensembles.

%% file: tables/ablations_side_by_side.tex
\begin{table}
\caption{\textbf{Left - variance and Mutual Information (MI)} show higher diversity for \modelname{} than for ensembles, while lower ECE shows that our approach yields better calibrated predictions.
All metrics are computed on DeepLabV3-ResNet50.
\textbf{Right - single-dilation:} OoD detection results (AUPR) for single dilation models (SD) with different dilation rates (1, 12, 24, 36), compared to standard multi-dilation \modelname{}. First row shows results for the global head, second row adds the probes, bottom two rows show the absolute and relative improvement
}
    \tiny
    \hspace{20px}
    \begin{minipage}{.4\linewidth}
    \begin{tabular}{l|ccc|ccc}
        \toprule
        \rule{0pt}{4px}
        & \multicolumn{3}{c|}{StreetHazards} & \multicolumn{3}{c}{BDD-Anomaly} \\
        \rule{0pt}{8px}
        Method & Var.$\uparrow$ & MI$\uparrow$ & ECE$\downarrow$ & Var.$\uparrow$ & MI$\uparrow$ & ECE$\downarrow$\\
        \midrule
        \midrule
       \globalhead & - & - & .038 & - & - & .123\\ 
       \multiheadens & 0.20 & .004 & .039 & 0.30 & .022 & .104 \\
        \deepens     & 0.54 & .012 & .032 & 1.19 & .054 & .103\\
        \midrule
        \modelname & 1.05 & .034 & .031 & 1.34 & .062 & .093 \\
        \bottomrule
    \end{tabular}
    \end{minipage}
    \hspace{35pt}
    \begin{minipage}{.4\linewidth}
    \begin{tabular}{l|rrrr|r}
        \toprule
                & \rotatebox{0}{SD1} & \rotatebox{0}{SD12} & \rotatebox{0}{SD24} & \rotatebox{0}{SD36} & \rotatebox{0}{MOoSe} \\
                \midrule
        Var.            & 0.41      & 1.05      & 0.93      & 0.83          & 1.34          \\
        \midrule
        \globalhead     & 8.2       & 8.1       & 9.1       & 8.6           & 10.2          \\
        +probes         & 10.1      & 10.1      & 10.6      & 10.5          & 13.1          \\
        \midrule
        Chng.          & 1.9        & 2.0       & 1.5       & 1.9           & 2.9           \\
        Chng. \%       & 22.7       & 24.9      & 16.4      & 22.3          & 27.9          \\
        \bottomrule
    \end{tabular}
    \end{minipage}
    \label{tab:ablations}
\end{table}

%% file: figures/graphs/foreground_noise.tex
\begin{figure}
     \centering
     \begin{subfigure}[b]{0.35\textwidth}
        \begin{tabular}[b]{@{}c@{}}
            \includegraphics[width=.5\linewidth]{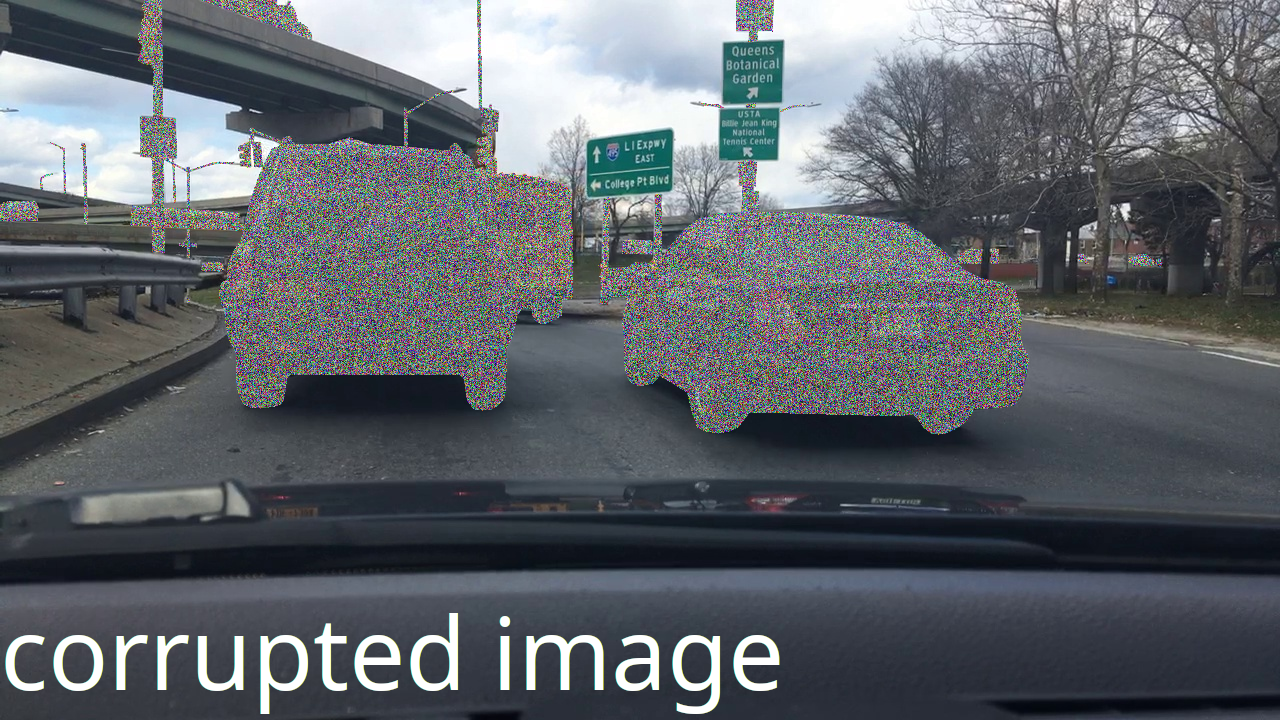}\hspace{5px}%
            \includegraphics[width=.5\linewidth]{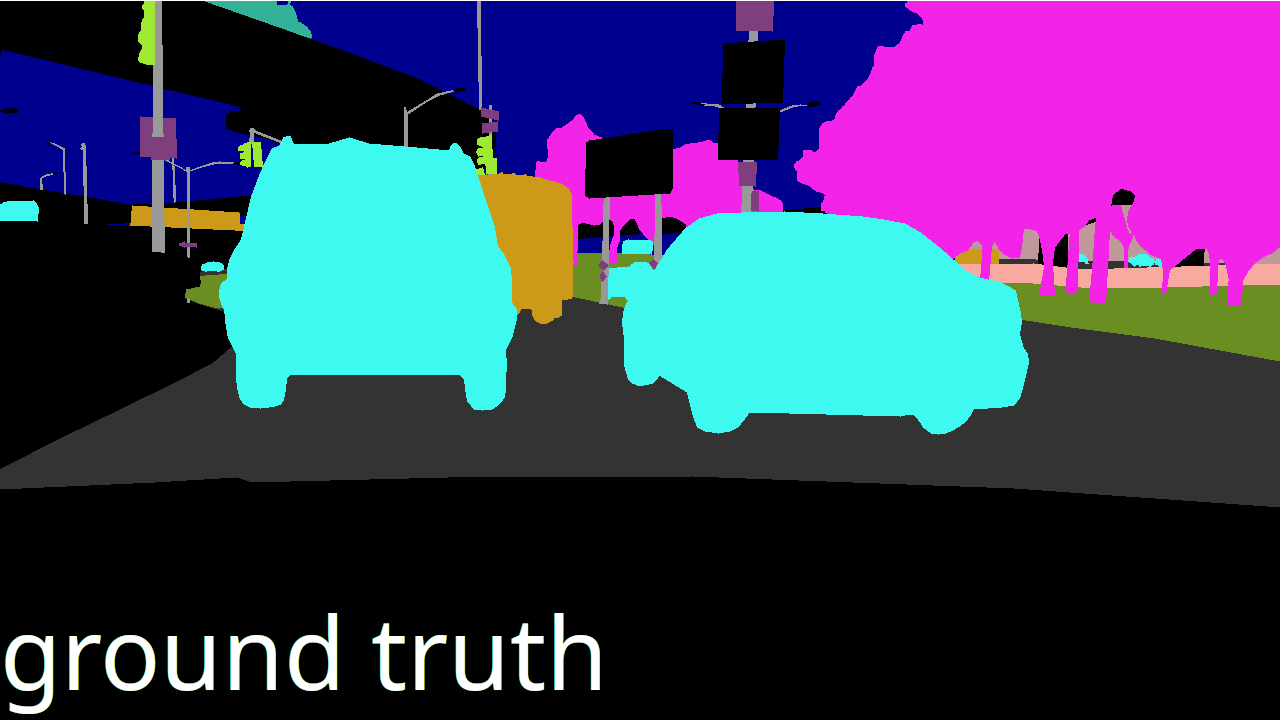}\\
            \includegraphics[width=.5\linewidth]{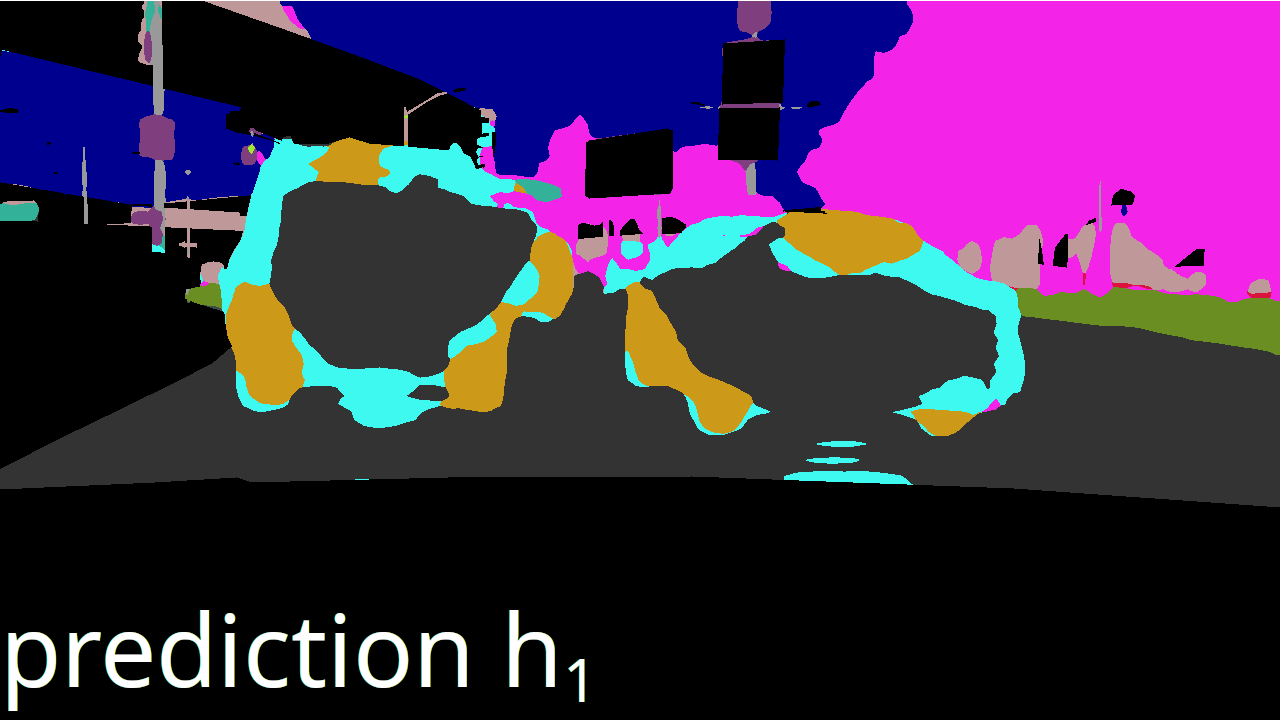}\hspace{5px}%
            \includegraphics[width=.5\linewidth]{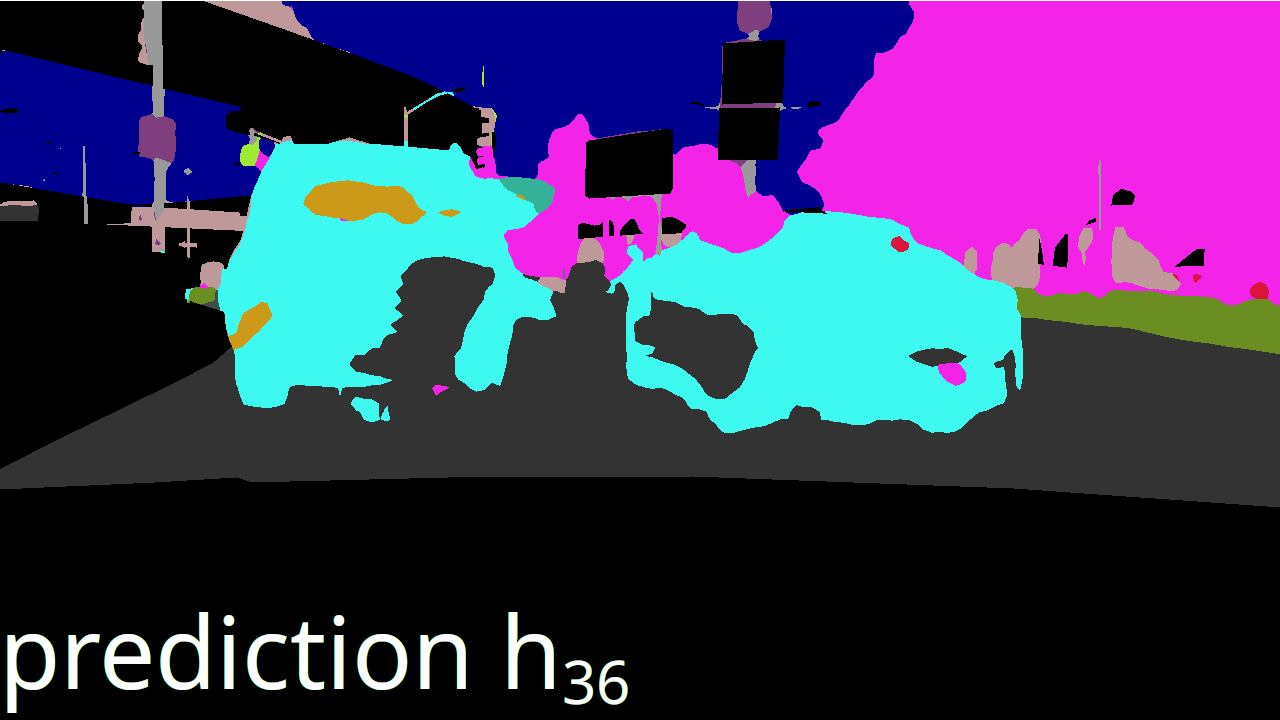}
        \end{tabular}%
        \vspace{9px}
    \end{subfigure}
    \hspace{15px}
    \begin{subfigure}[b]{0.33\textwidth}
         \centering
         \includegraphics[width=\textwidth]{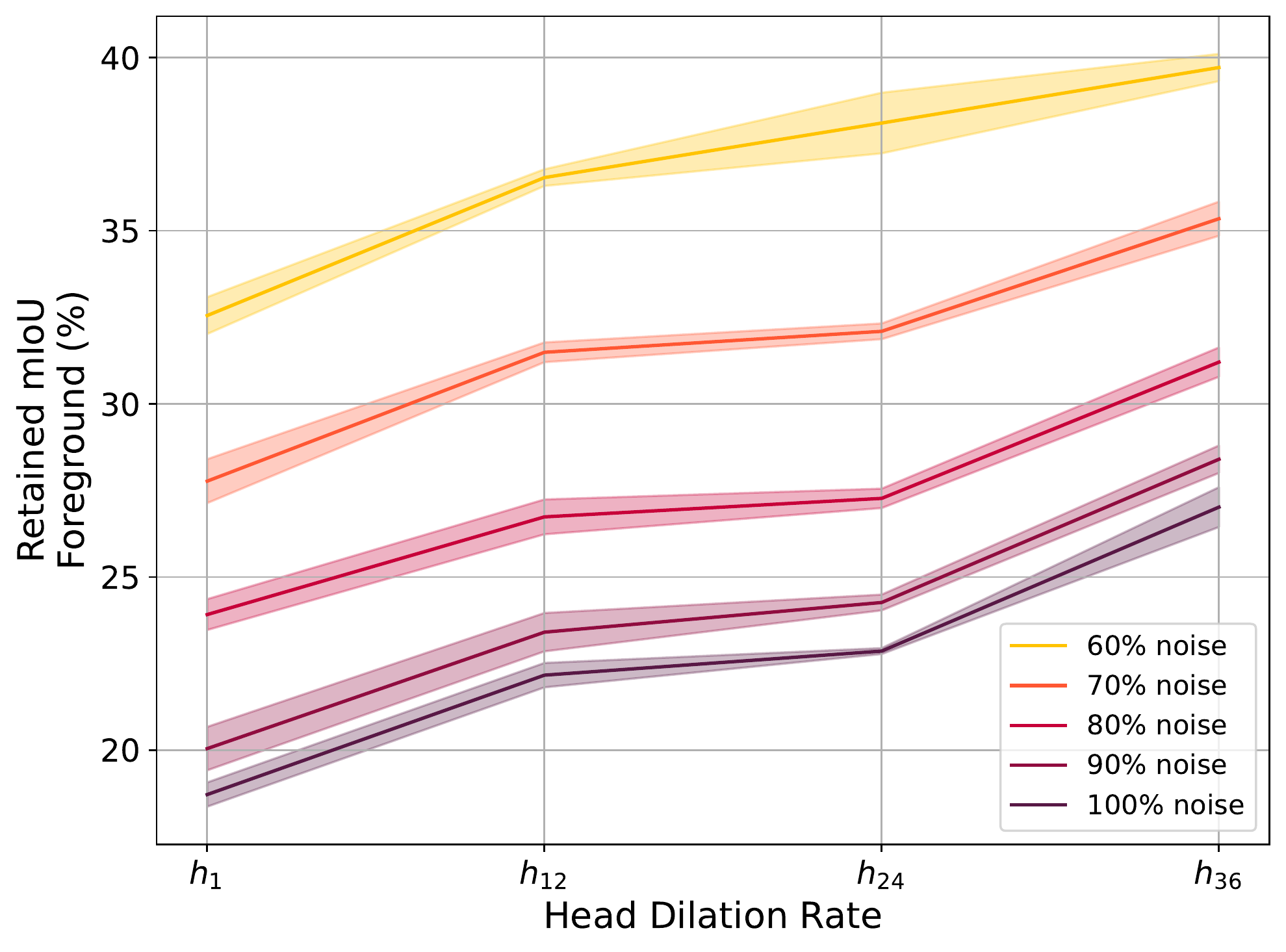}
     \end{subfigure}
    \caption{
    \textbf{Left - example of foreground corruption:} the cars are corrupted with noise and the head with the largest dilation rate ($h_{36}$) can still largely segment them, unlike the no-dilation head $h_{1}$.
    \textbf{Right - corruption robustness:} We evaluate semantic segmentation of each probe on the corrupted foreground objects. The retained mIoU ($\text{mIoU}_{corrupt}/\text{mIoU}_{clean}$) increases with dilation rate, indicating more reliance on context. Results for BDD-Anomaly on DeepLabV3.
    }
    \label{fig:fg_corruption}
\end{figure}


%% file: sections/6_conclusion.tex
In this work we proposed a simple and effective approach for improving dense out-of-distribution detection by leveraging the properties of segmentation decoders to obtain a set of diverse predictions. Our experiments showed that \modelname{} yields consistent gains on a variety of datasets and model architectures, and that it compares favorably with computationally much more expensive ensembles. We showed that our approach also outperforms other state-of-the-art approaches, and that due to its simplicity it could be easily combined with them.
Even though we tested our method on various architectures, and despite the versatility of the main idea, one current limitation of \modelname{} is its reliance on a specific architectural paradigm: the spatial pyramid. 
We also identified false positives among small objects to be an inherent failure mode of our approach, which potentially could be mitigated by combining \modelname{} with alternative concepts that act at a single contextual scale.